\documentclass[10pt,twocolumn,letterpaper]{article}

\usepackage{cvpr}      

\usepackage{color}
\usepackage{multirow}
\usepackage{makecell}
\usepackage{diagbox}
\newcommand{\NickName}{}
\usepackage{amsmath, amssymb, amsthm}
\usepackage{siunitx}
\usepackage{graphicx}
\usepackage{xcolor}
\usepackage{multicol}
\usepackage{colortbl}
\usepackage{tabularx}
\usepackage{makecell}
\usepackage{multirow}
\usepackage{diagbox}
\usepackage{booktabs}
\usepackage{wrapfig}
\usepackage{enumitem}
\usepackage{algorithm}
\usepackage{algpseudocode}
\usepackage[normalem]{ulem}
\usepackage{tgcursor}
\usepackage{bbm}
\usepackage[utf8]{inputenc}
\usepackage{amsmath}
\usepackage{amssymb}
\usepackage{marvosym}
\definecolor{cvprblue}{rgb}{0.21,0.49,0.74}
\definecolor{lightgray}{rgb}{0.7, 0.7, 0.7}
\definecolor{lightblue}{RGB}{230,240,255}
\definecolor{lightgreen}{RGB}{230,255,230}
\definecolor{lightyellow}{RGB}{255,255,230}
\definecolor{lightred}{RGB}{255,230,230}
\definecolor{lightlightgray}{gray}{.95}
\definecolor{lightlightblue}{RGB}{240,245,255}
\definecolor{lightlightgreen}{RGB}{240,255,240}
\definecolor{lightlightyellow}{RGB}{255,255,240}
\definecolor{lightlightred}{RGB}{255,240,240}
\definecolor{lightlightlightgray}{gray}{.99}
\definecolor{lightlightlightblue}{RGB}{247,250,255}
\definecolor{lightlightlightgreen}{RGB}{247,255,247}
\definecolor{lightlightlightyellow}{RGB}{255,255,247}
\definecolor{lightlightlightred}{RGB}{255,247,247}

\usepackage[pagebackref,breaklinks,colorlinks,allcolors=cvprblue]{hyperref}

\newcommand{\mymethod}{InfiniteVGGT}

\title{InfiniteVGGT: Visual Geometry Grounded Transformer for Endless Streams}

\author{
    Shuai Yuan$^{1}$ \quad Yantai Yang$^{1, 2}$ \quad Xiaotian Yang$^{1}$ \quad Xupeng Zhang$^{1}$ \\
    Zhonghao Zhao$^{1}$ \quad Lingming Zhang \quad Zhipeng Zhang$^{1}$\textsuperscript{\Letter}\\
    \\
    $^{1}$AutoLab, School of Artificial Intelligence, Shanghai Jiao Tong University \quad
    $^{2}$ Anyverse Dynamics
}

\begin{document}

\newcommand\blfootnote[1]{%
  \begingroup
  \renewcommand\thefootnote{}\footnote{#1}%
  \addtocounter{footnote}{-1}%
  \endgroup
}
\maketitle
\blfootnote{\textsuperscript{\Letter} Corresponding Author.}

\begin{abstract}
The grand vision of enabling persistent, large-scale 3D visual geometry understanding is shackled by the irreconcilable demands of scalability and long-term stability. While offline models like VGGT achieve inspiring geometry capability, their batch-based nature renders them irrelevant for live systems. Streaming architectures, though the intended solution for live operation, have proven inadequate. Existing methods either fail to support truly infinite-horizon inputs or suffer from catastrophic drift over long sequences. We shatter this long-standing dilemma with \textbf{\mymethod{}}, a causal visual geometry transformer that operationalizes the concept of a rolling memory through a bounded yet adaptive and perpetually expressive KV cache. Capitalizing on this, we devise a training-free, attention-agnostic pruning strategy that intelligently discards obsolete information, effectively ``rolling'' the memory forward with each new frame. Fully compatible with FlashAttention, \mymethod{} finally alleviates the compromise, enabling infinite-horizon streaming while outperforming existing streaming methods in long-term stability. The ultimate test for such a system is its performance over a truly infinite horizon, a capability that has been impossible to rigorously validate due to the lack of extremely long-term, continuous benchmarks. To address this critical gap, we introduce the \textbf{Long3D} benchmark, which, for the first time, enables a rigorous evaluation of continuous 3D geometry estimation on sequences about 10,000 frames. This provides the definitive evaluation platform for future research in long-term 3D geometry understanding. Code is available at: https://github.com/AutoLab-SAI-SJTU/InfiniteVGGT
\end{abstract}
    
\section{Introduction}
\label{sec:intro}

The dense reconstruction of 3D scenes from 2D images constitutes a cornerstone problem in geometric vision, serving as the bedrock for critical applications such as augmented reality (AR)~\cite{2024gpsgaussian,20243dclothed,2025mosca} and embodied AI~\cite{2024embodiedocc,20254DTAM, black2024pi_0,kim24openvla, yang2025efficientvla,li2024cogact,brohan2023rt2visionlanguageactionmodelstransfer}. Historically, the domain has been dominated by classical methods rooted in Structure-from-Motion (SfM)~\cite{2010Building,2011Building,2025Robust,2006PhotoTourism,2013LinearTime,2016SfM} and Multi-View Stereo (MVS)~\cite{2009Accurate,2020Cascade}. While capable of high-fidelity geometric optimization, these approaches are characterized by fragmented, multi-stage pipelines that are prohibitively slow, and prone to cascading errors. A paradigm shift has been catalyzed by the advent of end-to-end deep learning frameworks, which transcend these limitations by holistically inferring 3D structure from raw image data. Models such as DUSt3R~\cite{2024dust3r}, VGGT~\cite{2025vggt}, and their derivatives~\cite{2024mast3r,2025Fast3R} have reshaped the landscape, championing fully data-driven methodologies that achieve globally consistent reconstructions with unprecedented efficiency.

\begin{figure}[tb!]
    \centering
    \includegraphics[width=0.99\linewidth]{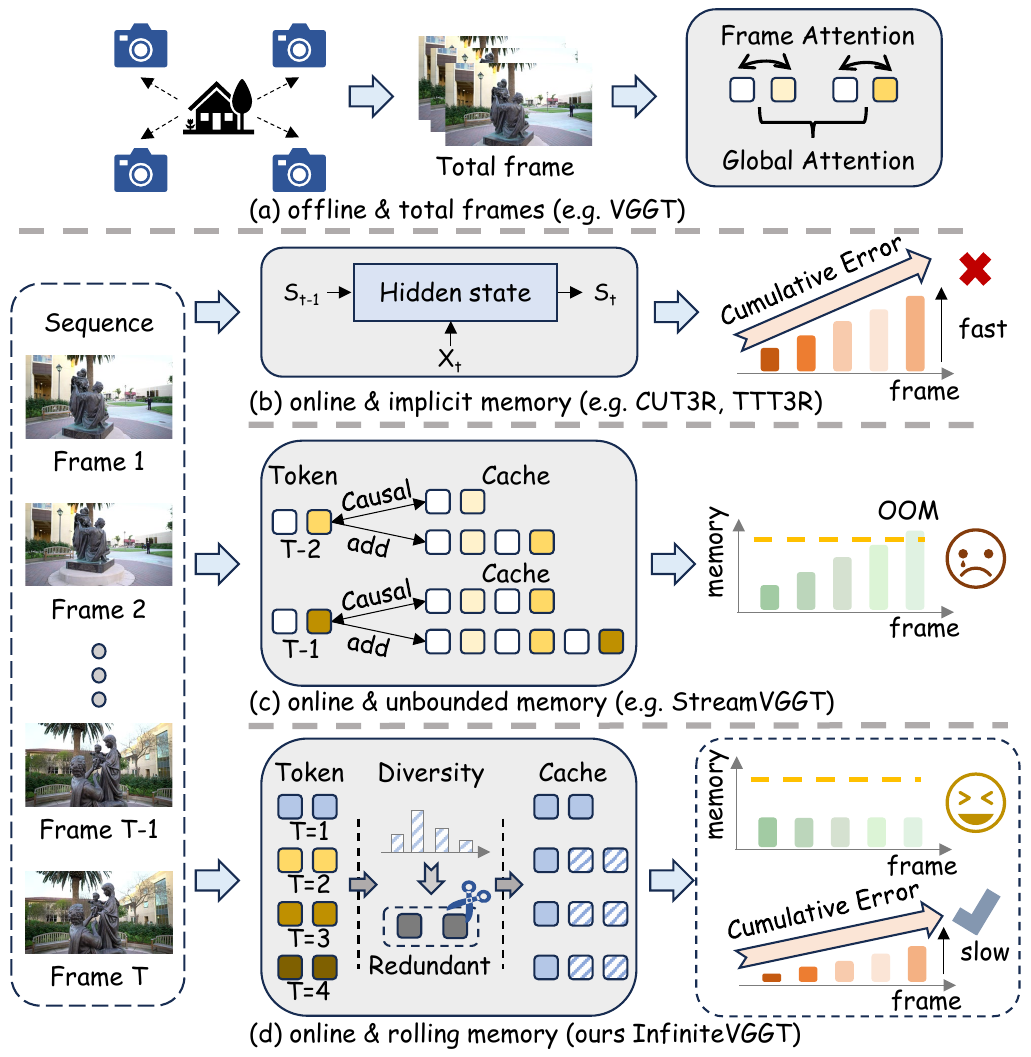}
    \caption{\textbf{Paradigm Comparison} between previous online and offline 3D geometry understanding and our \mymethod{}.}
    \label{fig:visualization_comparison}
    \vspace{-15pt}
\end{figure}

As these end-to-end models mature, the contemporary landscape has become defined by a fundamental dichotomy between offline batch processing~\cite{2025vggt,2025fastvggt,2025vggtlong} and online streaming paradigms~\cite{2025streamVGGT,2025stream3r,2025cut3r,2025ttt3r}. As illustrated in Fig.~\ref{fig:visualization_comparison}(a), offline methods masterfully exploit multi-view geometric constraints to achieve superior geometric fidelity, rendering them ideal for short-term reconstructions where data is fully pre-captured. This batch-centric paradigm, however, is fundamentally ill-suited for online applications or unbounded sequences due to its prohibitive GPU memory footprint~\cite{2025fastvggt}. Conversely, streaming architectures are conceptually tailored for online scenarios, such as robotics, by processing inputs sequentially to provide immediate perceptual feedback. Their theoretical appeal lies in handling infinite-length scene flows. Yet, this promise is largely unrealized in practice. One paradigm, namely explicit history accumulation frameworks like StreamVGGT~\cite{2025streamVGGT}, betrays its online intent by accumulating unbounded Key-Value (KV) stores (Fig.~\ref{fig:visualization_comparison}(c)), a path that inevitably leads to crippling memory and computational overheads. The other one, namely implicit state compression mechanisms, such as those in CUT3R~\cite{2025cut3r} and TTT3R~\cite{2025ttt3r} (Fig.~\ref{fig:visualization_comparison}(b)), make a Faustian bargain, where they compress history into a simple RNN hidden state to guarantee bounded resources, but in doing so, discard critical information, thereby exacerbating long-term drift and compromising robustness. Then, a question naturally arises that \textit{is it possible to selectively retain critical historical information to ensure temporal consistency, while still operating within the bounded resources required for a truly online system?}

The key to escaping this dilemma lies not in a more complex model, but in a pivotal insight into the nature of the data itself. We observe that in contiguous camera trajectories, minimal viewpoint shifts create massive token-level redundancy within the KV cache. This is not a trivial matter, as each frame adds approximately 1,000 tokens, the cache rapidly explodes to a scale ($\mathcal{O}(10^5)$ tokens within 100 frames) that necessitates hardware-optimized kernels like FlashAttention just to remain functional. Herein lies a fundamental paradox that these kernels achieve their speed by circumventing the materialization of the full $\mathcal{O}(N^2)$ attention matrix, yet traditional pruning methods rely on accessing these very weights to gauge token importance. Consequently, the tool required to manage the size of the cache prevents us from intelligently shrinking it. To resolve this impasse, we introduce an elegant solution by leveraging key cosine similarity as an efficient, attention-independent proxy for token importance. This allows us to identify and discard redundant tokens before the costly attention computation, thereby preserving the efficiency of optimized kernels while surgically shrinking the cache and finally paving the way for truly scalable streaming reconstruction.

Building upon this, we introduce \textbf{\mymethod{}}, which embodies a novel ``rolling memory'' paradigm for online 3D geometry understanding. It avoids the unbounded memory growth inherent in explicit history accumulation frameworks while simultaneously mitigating the information drift that plagues implicit state compression methods. Our rolling memory achieves this by continuously and dynamically refreshing its contents through a deeply integrated, multi-level retention strategy. At its foundation, the strategy abandons intuitive and coarse frame-wise deletion, selectively preserving individual tokens to maintain crucial long-term context. This granular process is then governed by a dynamic budget that is intelligently structured across the model's architecture. It functions layer-wise by assigning a unique token budget to each layer, resulting in layer-specific and specialized KV caches. This systematic control system operates without materializing attention weights, ensuring full compatibility with FlashAttention, and ultimately enables a system with a strictly bounded GPU memory footprint capable of processing infinite sequences.

The ultimate test for such a system is its performance over a truly infinite horizon, a capability that has been impossible to rigorously validate due to the lack of continuous, long-term benchmarks. To address this gap, we introduce the \textbf{Long3D} benchmark, which, for the first time, enables a rigorous evaluation of continuous 3D geometry estimation on sequences about 10,000 frames. This provides the definitive evaluation platform for future research in long-term 3D scene understanding and reconstruction. 



Our contributions are threefold: $\spadesuit$ An unbounded memory architecture \mymethod{} for continuous 3D geometry understanding, built on a novel, dynamic, and interpretable explicit memory system.
$\spadesuit$ State-of-the-art performance on long-sequence benchmarks and a unique capability for robust, infinite-horizon reconstruction without memory overflow.
$\spadesuit$ The Long3D benchmark, a new dataset for the rigorous evaluation of long-term performance, addressing a critical gap in the field.
\section{Related Work}
\label{sec:Related work}

\paragraph{Classical Offline and Online Reconstruction.}
Traditional 3D vision methods fall into two primary paradigms distinguished by their operational constraints, namely offline batch processing and online streaming. The cornerstone of offline reconstruction is Structure-from-Motion (SfM). SfM pipelines~\cite{2010Building,2011Building,2025Robust,2006PhotoTourism,2013LinearTime,2016SfM}, epitomized by COLMAP~\cite{2016SfM}, perform a global Bundle Adjustment (BA)~\cite{2000Multiple} across all views and points to achieve maximum global accuracy. While computationally demanding, this batch optimization produces highly precise results that subsequently serve as a foundation for Multi-View Stereo (MVS)~\cite{2015MultiView,2016Pixelwise,2023AdaptivePatch,2015Massively} algorithms to generate dense models. In stark contrast, online streaming methods, prominently represented by Simultaneous Localization and Mapping (SLAM), prioritize online performance. These systems incrementally estimate the camera trajectory, employing a range of techniques that include feature-based~\cite{2015orbslam}, direct~\cite{2014lsd-slam}, and dense~\cite{2011DTAM} approaches.

\paragraph{Learning-based Offline 3D Reconstruction.}
Recent advances in offline 3D reconstruction have seen classical multi-stage pipelines give way to unified, feed-forward architectures. Early works like DUSt3R~\cite{2024dust3r} and MASt3R~\cite{2024mast3r} formulate reconstruction as a pairwise pointmap regression problem, imposing a computationally expensive global alignment stage to aggregate multi-view information. VGGT~\cite{2025vggt} addresses this by introducing a large transformer that jointly predicts camera poses, depth, and feature tracks in a single forward pass. More recently, $\pi^3$~\cite{2025pi3} refines VGGT to operate independently of a fixed reference frame. However, the input length of such large models remains a bottleneck. To extend scalability, VGGT-Long~\cite{2025vggtlong} decomposes long trajectories into sub-maps at the cost of single-pass simplicity. In a different approach, Sail-Recon~\cite{2025sailrecon} enhances scene regression using a subset of anchor images to create a global neural representation for efficient localization of all other images. Focusing instead on computational efficiency, FastVGGT~\cite{2025fastvggt} accelerates the forward process through a training-free token merge mechanism. This method exploits attention redundancy to preserve key geometric cues, achieving a 4$\times$ speedup on 1000-frame sequences while reducing drift.

\noindent \textbf{Learning-based Online 3D Reconstruction.}
The early transformer-based methods, such as Spann3R~\cite{2024spann3r} and Point3R~\cite{2025point3r}, pioneered online forward-pass reconstruction using explicit spatial or pointer memory. This paradigm was refined by StreamVGGT~\cite{2025streamVGGT} and Stream3R~\cite{2025stream3r}, which apply causal attention and a KV cache to process sequences on-the-fly. However, the reliance on an ever-growing KV cache leads to prohibitive increases in memory and computation, rendering these models impractical for truly long streaming inputs. To circumvent this scaling issue, WinT3R~\cite{2025wint3r} employs a sliding-window mechanism to balance reconstruction quality with latency. This design inherently limits the temporal receptive field and can cause drift. Although WinT3R attempts to mitigate this with a global camera-token pool, it still falls short of supporting infinite-length reconstruction. Seeking to overcome these limitations, another line of research adopts RNN-based architectures. CUT3R~\cite{2025cut3r}, for example, uses continuously updated states to accommodate arbitrary-length image streams. Building on this foundation, TTT3R~\cite{2025ttt3r} introduces test-time training rules to improve length generalization, enabling the online processing of thousands of frames. Nevertheless, catastrophic forgetting caused by transitionally compressed memory remains a fundamental challenge, limiting the ability of capturing long-range temporal dependencies. To mitigate these limitations, we propose an online streaming framework capable of infinite-length 3D geometry reconstruction by introducing a hierarchical, dynamic rolling memory that preserves long-term dependencies to reduce both drift and catastrophic forgetting.

\section{Method}
\label{sec:Method}

\subsection{Preliminaries}
\paragraph{From Offline to Online 3D Reconstruction.}
The offline model VGGT~\cite{2025vggt} processes a batch of $N$ images $\{ I_i \in \mathbb{R}^{H \times W \times 3} \}_{i=1}^N$ in a single forward pass. It alternately applies frame ($\mathcal{F_\theta}$) and global ($\mathcal{G_\theta}$) interaction across 24 self-attention layers to jointly estimate a set of 3D quantities,
\begin{equation}
    (\mathbf{g}_i, D_i, P_i, T_i)_{i=1}^N = \phi(\mathcal{F_\theta}\left( \{I_i\}_{i=1}^N \right),\mathcal{G_\theta}\left( \{I_i\}_{i=1}^N \right)),
\end{equation}
where  $\mathbf{g}_i\in \mathbb{R}^{9}$ represents the camera parameters, $D_i \in \mathbb{R}^{H \times W}$ is the depth map, $P_i \in \mathbb{R}^{H \times W \times 3}$ is the point map, and $T_i \in \mathbb{R}^{H \times W \times C}$ are point-tracking features.

To adapt this architecture for online and streaming usage, models like StreamVGGT~\cite{2025streamVGGT} substitute the global interaction $\mathcal{G_\theta}$ with a causal temporal attention module $\mathcal{T_\theta}$. This allows the model to process frames incrementally. At any given timestep $t$, the module generates the output for the current frame $I_t$ by leveraging a KV cache, $\mathcal{C}_{t-1}$, that stores the context from all previous frames,
\begin{equation}
    (\mathbf{g}_t, D_t, P_t, T_t) = \phi(\mathcal{F_\theta}\left( I_t\right),\mathcal{T_\theta}( I_t, \mathcal{C}_{t-1}))
\end{equation}
The KV cache $\mathcal{C}_t = \left\{ (\mathcal{K}_t^{(l)}, \mathcal{V}_t^{(l)}) \right\}_{l=1}^{N_L}$, where $N_L$ is the total number of causal attention layers, is contiguously updated by combining the new keys and values. 
The cache size grows linearly ($\mathcal{O}(t)$) with the sequence length $t$.

\begin{figure*}[!t]
    \centering
    \vspace{-1em}
    \includegraphics[width=\textwidth]{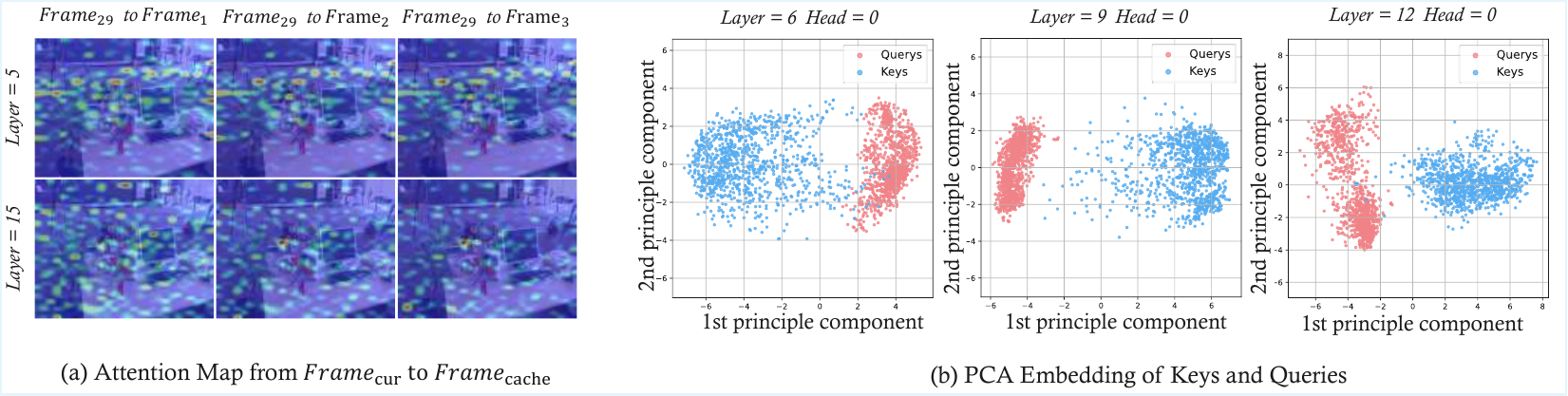}
    \caption{\textbf{Visualization Results.} \textbf{(a)} Attention maps from the current frame to adjacent historical cached frames, demonstrating near-identical distributions due to minimal viewpoint shifts in online streaming camera motion. \textbf{(b)} PCA embeddings of query (Q) and key (K) vectors for representative layers and heads, revealing clustering and redundancy in the feature space.}
    \label{fig:visual}
    \vspace{-1em}
\end{figure*}  

\subsection{Motivation and Analysis}
 As previously discussed, the KV cache, which functions as explicit memory, grows linearly with each new frame, leading to unsustainable memory demands over time. The central challenge is thus to maintain a fixed-size \textit{rolling memory}. This requires an intelligent eviction strategy that preserves valuable information while discarding redundancy.

\noindent \textbf{Are Attention Scores a Feasible Eviction Criterion?}
An intuitive approach is to use attention scores as a proxy for token importance. 
This idea is initially compelling because sequential input frames in 3D reconstruction possess high spatio-temporal redundancy due to significant viewpoint overlap. 
We empirically validate this by extracting the patch-embedded tokens from the backbone of StreamVGGT~\cite{2025streamVGGT} for adjacent frames, finding their cosine similarity consistently exceed 0.95. 
This high similarity stems from the DINO~\cite{2024DINOv2} backbone being trained as a semantic encoder with high invariance to slight changes in viewpoint. 
It prioritizes ``what" is seen over ``from where" it is seen, which further confirms the extensive redundancy present in the tokens fed into the subsequent aggregator module.
As a result, the query frame $I_t$ often assigns near-identical attention weights to historical frames that share a similar perspective (Fig.~\ref{fig:visual} (a)). 
This observation suggests that an attention-based eviction strategy could effectively prune the cache. 
However, this approach introduces a critical computational dilemma. 
The KV cache in these architectures must scale to hundreds of thousands or even millions of tokens ($|\mathcal{C}_t| \gg 10^5$). To manage this scale online, the causal attention mechanism ($\mathcal{T_\theta}$) fundamentally depends on hardware-optimized kernels, such as FlashAttention, which mitigate memory bandwidth bottlenecks by never explicitly materializing the full attention matrix. 
This reliance creates an irreconcilable conflict that any token filtering strategy predicated on attention scores requires the materialization of the full attention weight matrix, which is the very operation that optimized kernels are designed to bypass. 
Executing this operation would be computationally prohibitive and would negate the low-latency inference essential for streaming systems. We therefore argue that this paradigm is suboptimal and motivate the need for an alternative approach to construct an efficient rolling memory.

\noindent \textbf{Key Diversity as a Redundancy Proxy.}
Instead of estimating token salience through attention weights, we measure redundancy in the key space. 
As illustrated in Fig.~\ref{fig:visual}(b), PCA visualizations of the key and query spaces reveal that queries ($\mathbf{q}_t$) from the current frame and cached key vectors ($\mathbf{k}_{t-1}$) consistently occupy distinct, nearly orthogonal subspaces across layers. This geometric separation persists over time, confirming that key-space similarity provides a stable measure of redundancy.
Therefore, distinct keys will be more aligned with the query, and in turn, can be preserved as more salient keys.
Building on this, we define the negative cosine similarity as diversity score to quantify this dispersion, hypothesizing that keys are the principal components and provide the most effective mechanism for quantifying redundancy.
This metric efficiently captures the dispersion of key representations in feature space and is independent of the current query.
Tokens with higher diversity scores correspond to those most dissimilar from the global mean, and are thus retained during cache compression. 
As a result, the cache preserves the most informative subset of tokens while maintaining a minimal memory footprint.

\begin{figure*}[t]
    \centering
    \vspace{-1em}
    \includegraphics[width=\textwidth]{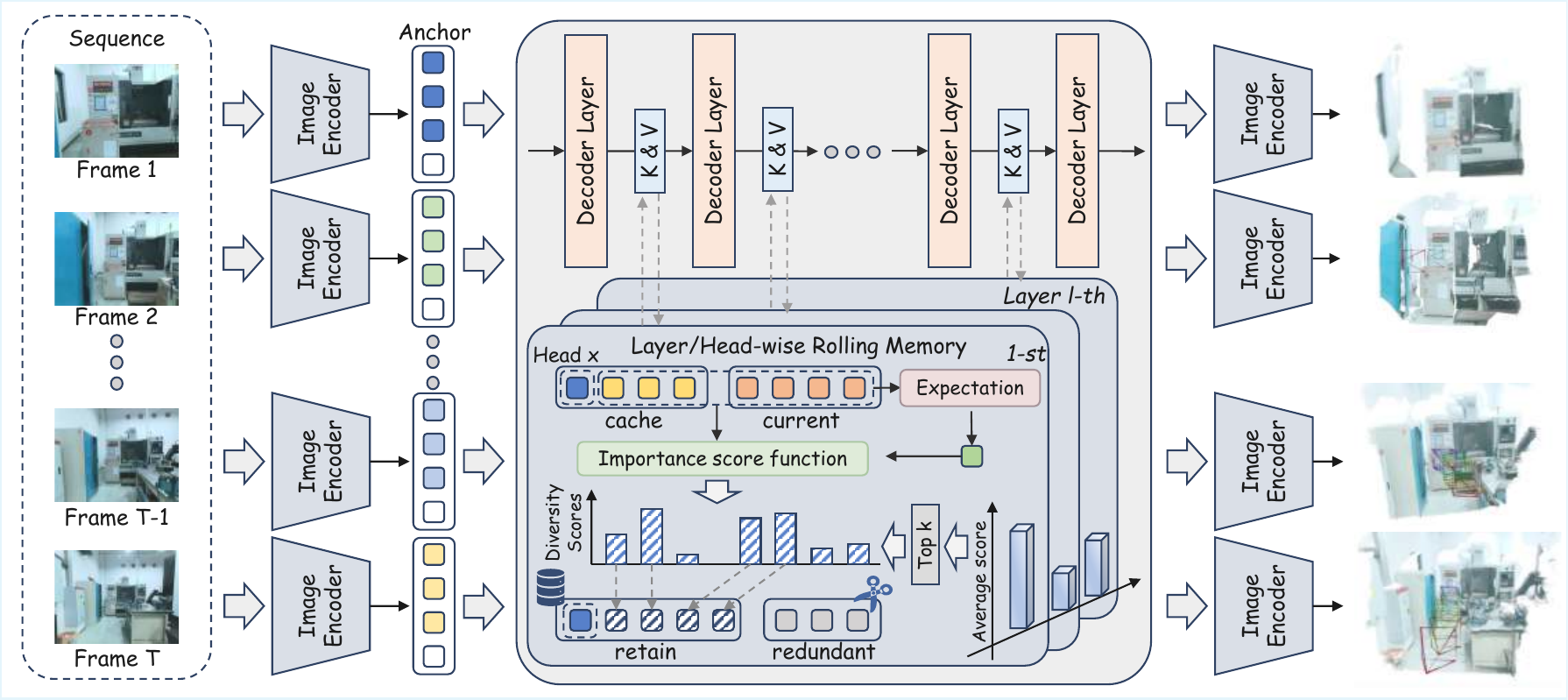}
    
    \caption{\textbf{Overview of the InfiniteVGGT}, illustrating a rolling memory paradigm that prunes KV cache contents to prevent VRAM accumulation over time, employing key cosine similarity and adaptive layer-wise allocation for 3D geometry understanding.}
    \label{fig:pipeline}
    \vspace{-1em}
\end{figure*}
\subsection{Diversity-aware Rolling Memory}
\paragraph{Immutable Anchor Token.}
As illurstrated in Fig.~\ref{fig:pipeline}, our rolling memory pipeline commences by establishing an immutable set of \emph{anchor tokens}, defined as the complete KV cache derived from the initial input frame. This design choice is motivated by the architectural foundation of VGGT~\cite{2025vggt}, wherein all subsequent 3D predictions are rigidly aligned to the coordinate system of the first frame, which serves as the canonical global reference. Any alteration or pruning of these initial tokens would irreversibly compromise geometric consistency across the entire reconstruction. Accordingly, we designate the first-frame cache as the immutable anchor set $\mathcal{C}^{(l,h)}_{\text{anc}}$ and exclude it from all subsequent compression operations.
For any given layer $l$ and head $h$, the total cache $\mathcal{C}^{(l,h)}_t$ is thus partitioned into the anchor set and a mutable candidate set, $\mathcal{C}^{(l,h)}_{t,cand}$, which contains all tokens from $t=2$ onwards. 

\paragraph{Diversity-quantified Token Retention.}
Then, we apply our retention strategy $\mathcal{\pi}$ exclusively to the candidate set $\mathcal{C}^{(l,h)}_{t,cand}$ to retain the most informative tokens. 
This process is performed independently for each layer $l$ and head $h$ to account for their heterogeneous redundancy profiles. Our strategy begins by establishing a reference vector for each head's key space. This is achieved by computing the mean key $\mathbf{\mu}^{(l,h)}$. To ensure this metric captures directional variance exclusively, we operate on L2-normalized keys, where $\hat{\mathbf{k}}_i = \mathbf{k}_i / ||\mathbf{k}i||$. The mean key is thus the expectation over the set of normalized candidate keys $\hat{\mathcal{K}}_{t,cand}^{(l,h)}$, 
\begin{equation}
    \mathbf{\mu}^{(l,h)} = \mathbb{E}_{\mathbf{\hat{k}} \in \hat{\mathcal{K}}_{t,cand}^{(l,h)}}[\mathbf{\hat{k}}]
\end{equation}  
Next, we define a diversity score $s_{div}$ for each individual key $\hat{\mathbf{k}}_i$ to quantify its dissimilarity from this mean vector. Based on our previous analysis, we employ the negative cosine similarity as our metric,
\begin{equation}
    s_{div}^{(l,h)}(\hat{\mathbf{k}_i}) = - \operatorname{CosSim}(\mathbf{\mu}^{(l,h)}, \hat{\mathbf{k}}_i)
\end{equation}
This formulation ensures that keys with the lowest cosine similarity to the mean, which represent the most geometrically distinct features, are assigned the highest scores. Consequently, a high $s_{div}$ score signifies high informational salience, guiding the retention of the most valuable tokens.

\subsection{Layer-wise Adaptive Budget Allocation}
To optimize the KV cache, we introduce an adaptive, layer-wise budget allocation mechanism that assigns a non-uniform storage budget to each layer in proportion to its measured information diversity. 
This strategy is motivated by the observation that informational diversity is unevenly distributed across the model. 
Our analysis reveals that shallow layers, which amplify subtle inter-frame differences for spatial reasoning, exhibit high diversity. In contrast, both the initial layer, processing low-level statistics like color and brightness, and the deep layers, where representations converge towards a holistic semantic understanding, demonstrate significantly less diversity. To implement this principle, we first define a layer-wise average diversity score $s^{l}_{div}$ as the mean of all token diversity $s^{(l,h)}_{div}$ within that layer. The budget proportion $p^{l}_{bud}$ for each layer is then calculated via a softmax normalization of these scores,
\begin{equation}
    p^{l}_{bud} = \frac{\exp(s^l_{div} / \tau)}{\sum_{j=1}^L \exp(s^j_{div} / \tau)}
\end{equation}    
where $\tau$ is a temperature hyperparameter. 
The total budget for layer $l$ is $B^l = p_{bud}^l \cdot B_{total}$.
This budget $B^{(l,h)}$ is then enforced via a $\text{TopK}$ selection.
The final compressed cache $\tilde{\mathcal{C}}_t$ is the union of all retained candidate tokens $\tilde{\mathcal{C}}_{t,cand}^{(l,h)}$ and the immutable anchor set $\mathcal{C}_{anc}$.

\begin{figure}[t]
    \centering
    \includegraphics[width=\columnwidth]{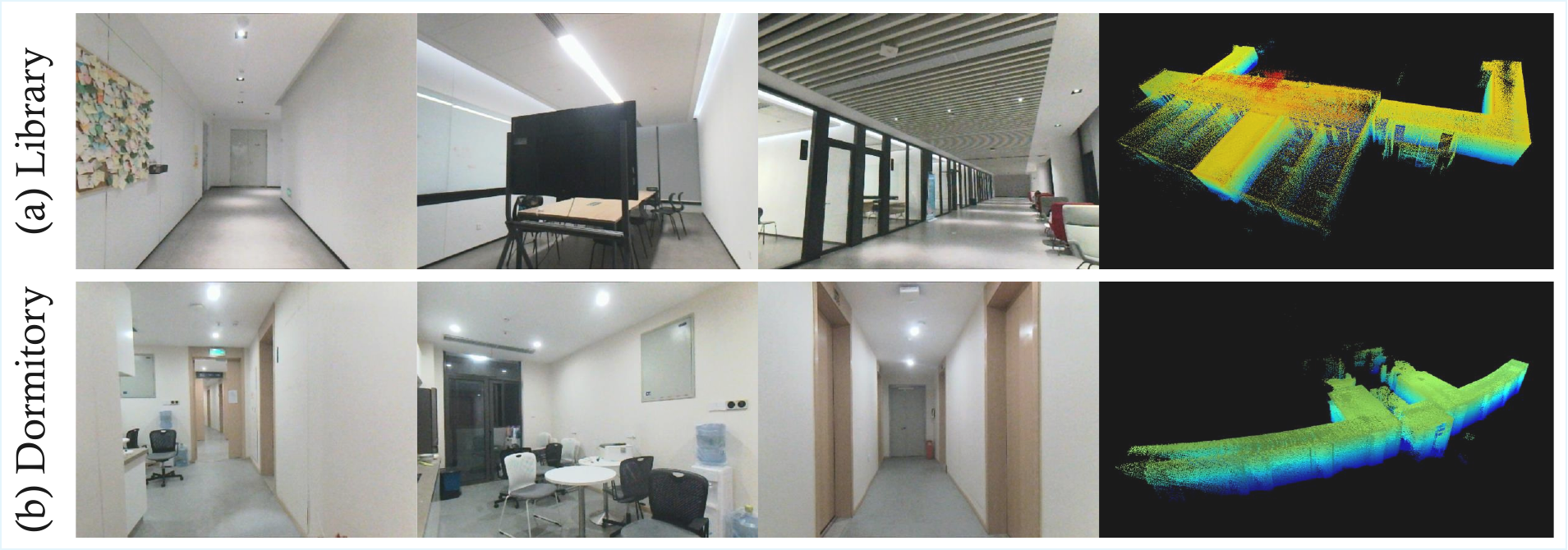}
    
    \caption{\textbf{Long3D Examples}. Views and global point clouds of different scenes.}
    \label{fig:long3d}
    \vspace{-1.5em}
\end{figure}

\begin{table*}[t]
\centering
\caption{
    \textbf{3D Reconstruction Results on 7-Scenes~\cite{20137scenes} and NRGBD~\cite{2022NeuralRGBD}.}
}
\vspace{-1em}
\resizebox{\textwidth}{!}{
\begin{tabular}{lcccccccccccccc}
    \toprule[0.17em]
    {\multirow{4}{*}{\textbf{Method}}} &
    {\multirow{4}{*}{\textbf{Input}}} &
    \multicolumn{6}{c}{\textbf{7-Scenes}} &
    \multicolumn{6}{c}{\textbf{NRGBD}} \\
    \cmidrule(r){3-8} \cmidrule(r){9-14}
    & &
    \multicolumn{2}{c}{Acc. $\downarrow$} &
    \multicolumn{2}{c}{Comp. $\downarrow$} &
    \multicolumn{2}{c}{NC $\uparrow$} &
    \multicolumn{2}{c}{Acc. $\downarrow$} &
    \multicolumn{2}{c}{Comp. $\downarrow$} &
    \multicolumn{2}{c}{NC $\uparrow$} & \\
    \cmidrule(r){3-4} \cmidrule(r){5-6} \cmidrule(r){7-8}
    \cmidrule(r){9-10} \cmidrule(r){11-12} \cmidrule(r){13-14}
    & &
    Mean & Med. &
    Mean & Med. &
    Mean & Med. &
    Mean & Med. &
    Mean & Med. &
    Mean & Med. & \\
    \midrule[0.08em]
    VGGT \textit{(Offline)}~\cite{2025vggt} & \multirow{5}{*}{\textit{300}} & \textcolor{lightgray}{\textit{OOM}} & \textcolor{lightgray}{\textit{OOM}} & \textcolor{lightgray}{\textit{OOM}} & \textcolor{lightgray}{\textit{OOM}} & \textcolor{lightgray}{\textit{OOM}} & \textcolor{lightgray}{\textit{OOM}} & \textcolor{lightgray}{\textit{OOM}} & \textcolor{lightgray}{\textit{OOM}} & \textcolor{lightgray}{\textit{OOM}} & \textcolor{lightgray}{\textit{OOM}} & \textcolor{lightgray}{\textit{OOM}} & \textcolor{lightgray}{\textit{OOM}} \\
    StreamVGGT~\cite{2025streamVGGT} && \textcolor{lightgray}{\textit{OOM}} & \textcolor{lightgray}{\textit{OOM}} & \textcolor{lightgray}{\textit{OOM}} & \textcolor{lightgray}{\textit{OOM}} & \textcolor{lightgray}{\textit{OOM}} & \textcolor{lightgray}{\textit{OOM}} & \textcolor{lightgray}{\textit{OOM}} & \textcolor{lightgray}{\textit{OOM}} & \textcolor{lightgray}{\textit{OOM}} & \textcolor{lightgray}{\textit{OOM}} & \textcolor{lightgray}{\textit{OOM}} & \textcolor{lightgray}{\textit{OOM}} \\
    CUT3R~\cite{2025cut3r} && 0.135 & 0.091 & 0.071 & 0.032 & 0.543 & 0.562 & 0.224 & 0.126 & 0.074 & 0.012 & 0.579 & 0.624 \\
    Point3R~\cite{2025point3r} && 0.047 & 0.027 & 0.029 & 0.011 & 0.563 & 0.596 & 0.076 & 0.043 & 0.014 & 0.005 & 0.618 & 0.695 \\
    TTT3R~\cite{2025ttt3r} && 0.041 & 0.025 & \textbf{0.024} & 0.005 & 0.565 & 0.599 & 0.103 & 0.045 & 0.025 & 0.005 & 0.608 & 0.673 \\
    \rowcolor{lightblue} \textbf{\NickName \mymethod{}} && \textbf{0.040} & \textbf{0.015} & 0.025 & \textbf{0.005} & \textbf{0.570} & \textbf{0.607} & \textbf{0.051} & \textbf{0.032} & \textbf{0.022} & \textbf{0.005} & \textbf{0.649} & \textbf{0.756}\\
    \midrule[0.08em]
    VGGT \textit{(Offline)}~\cite{2025vggt} & \multirow{5}{*}{\textit{400}} & \textcolor{lightgray}{\textit{OOM}} & \textcolor{lightgray}{\textit{OOM}} & \textcolor{lightgray}{\textit{OOM}} & \textcolor{lightgray}{\textit{OOM}} & \textcolor{lightgray}{\textit{OOM}} & \textcolor{lightgray}{\textit{OOM}} & \textcolor{lightgray}{\textit{OOM}} & \textcolor{lightgray}{\textit{OOM}} & \textcolor{lightgray}{\textit{OOM}} & \textcolor{lightgray}{\textit{OOM}} & \textcolor{lightgray}{\textit{OOM}} & \textcolor{lightgray}{\textit{OOM}} \\
    StreamVGGT~\cite{2025streamVGGT} && \textcolor{lightgray}{\textit{OOM}} & \textcolor{lightgray}{\textit{OOM}} & \textcolor{lightgray}{\textit{OOM}} & \textcolor{lightgray}{\textit{OOM}} & \textcolor{lightgray}{\textit{OOM}} & \textcolor{lightgray}{\textit{OOM}} & \textcolor{lightgray}{\textit{OOM}} & \textcolor{lightgray}{\textit{OOM}} & \textcolor{lightgray}{\textit{OOM}} & \textcolor{lightgray}{\textit{OOM}} & \textcolor{lightgray}{\textit{OOM}} & \textcolor{lightgray}{\textit{OOM}} \\
    CUT3R~\cite{2025cut3r} && 0.162 & 0.114 & 0.093 & 0.050 & 0.532 & 0.546 & 0.315 & 0.215 & 0.101 & 0.032 & 0.551 & 0.572 \\
    Point3R~\cite{2025point3r} && 0.049 & 0.023 & 0.026 & 0.009 & 0.559 & 0.589 & 0.093 & 0.045 & \textbf{0.024} & 0.005 & 0.613 & 0.685 \\
    TTT3R~\cite{2025ttt3r} && 0.052 & 0.031 & 0.027 & 0.005 & 0.558 & 0.587 & 0.140 & 0.070 & 0.058 & 0.014 & 0.599 & 0.657 \\
    \rowcolor{lightblue} \textbf{\NickName \mymethod{}} && \textbf{0.043} & \textbf{0.016} & \textbf{0.026} & \textbf{0.005} & \textbf{0.565} & \textbf{0.599} & \textbf{0.069} & \textbf{0.040} & 0.034 & \textbf{0.005} & \textbf{0.653} & \textbf{0.763}\\
    \midrule[0.08em]
    VGGT \textit{(Offline)}~\cite{2025vggt} & \multirow{5}{*}{\textit{500}} & \textcolor{lightgray}{\textit{OOM}} & \textcolor{lightgray}{\textit{OOM}} & \textcolor{lightgray}{\textit{OOM}} & \textcolor{lightgray}{\textit{OOM}} & \textcolor{lightgray}{\textit{OOM}} & \textcolor{lightgray}{\textit{OOM}} & \textcolor{lightgray}{\textit{OOM}} & \textcolor{lightgray}{\textit{OOM}} & \textcolor{lightgray}{\textit{OOM}} & \textcolor{lightgray}{\textit{OOM}} & \textcolor{lightgray}{\textit{OOM}} & \textcolor{lightgray}{\textit{OOM}} \\
    StreamVGGT~\cite{2025streamVGGT} && \textcolor{lightgray}{\textit{OOM}} & \textcolor{lightgray}{\textit{OOM}} & \textcolor{lightgray}{\textit{OOM}} & \textcolor{lightgray}{\textit{OOM}} & \textcolor{lightgray}{\textit{OOM}} & \textcolor{lightgray}{\textit{OOM}} & \textcolor{lightgray}{\textit{OOM}} & \textcolor{lightgray}{\textit{OOM}} & \textcolor{lightgray}{\textit{OOM}} & \textcolor{lightgray}{\textit{OOM}} & \textcolor{lightgray}{\textit{OOM}} & \textcolor{lightgray}{\textit{OOM}} \\
    CUT3R~\cite{2025cut3r} && 0.183 & 0.130 & 0.091 & 0.033 & 0.530 & 0.543 & 0.326 & 0.243 & 0.132 & 0.042 & 0.556 & 0.582 \\
    Point3R~\cite{2025point3r} && 0.063 & 0.026 & 0.031 & 0.015 & 0.555 & 0.583 & 0.113 & \textbf{0.048} & 0.037 & \textbf{0.005} & 0.613 & 0.684 \\
    TTT3R~\cite{2025ttt3r} && 0.062 & 0.036 & 0.029 & 0.005 & 0.552 & 0.577 & 0.165 & 0.084 & 0.095 & 0.015 & 0.594 & 0.648 \\
    \rowcolor{lightblue} \textbf{\NickName \mymethod{}} && \textbf{0.043} & \textbf{0.018} & \textbf{0.025} & \textbf{0.005} & \textbf{0.561} & \textbf{0.593} & \textbf{0.080} & 0.054 & \textbf{0.037} & 0.008 & \textbf{0.643} & \textbf{0.746} \\
    \bottomrule[0.17em]
\end{tabular}
}
\vspace{-1em}
\label{tab:mv_recon_long}
\end{table*}
\section{Long3D Benchmark}
\label{sec:Benchmark}
To address the critical lack of benchmarks for evaluating continuous, long-term 3D geometry estimation, we propose \textbf{Long3D}. 
Prior to this work, rigorously assessing a system's performance over extended, uninterrupted periods was infeasible, as existing benchmarks are either restricted to short sequences ($\leq 1000$ frames) or, like the 7-Scenes~\cite{20137scenes}, are merely collections of discontinuous clips, which prevents a proper assessment of long-term, uninterrupted performance. Long3D fills this critical void by providing the first framework for evaluating model robustness on truly continuous video streams. In total, our dataset features 5 challenging sequences captured in diverse indoor and outdoor environments, with each individual sequence ranging from about 2,000 to 10,000 frames. Fig.~\ref{fig:long3d} shows an example of our benchmark. This data was collected using a handheld 3D spatial scanner equipped with an IMU, a 3D LiDAR (360° horizontal by 59° vertical FOV), and an RGB camera ($800\times600$ at 10 Hz, 90° FOV). For each scene, the data consists of a global ground-truth point cloud and the corresponding uninterrupted sequence of RGB images. On our benchmark, we evaluate dense-view streaming reconstruction, where models process the entire image stream to generate a complete global point cloud. For evaluation, predicted and ground-truth point clouds are aligned using the Iterative Closest Point (ICP) algorithm, consistent with prior methods~\cite{2025cut3r,2025streamVGGT}. Performance is quantified using three established metrics, including Accuracy (Acc.), Completion (Comp.), Chamfer Distance(CD) and Normal Consistency (NC).

\section{Experiments}
\label{sec:Experiments}



\subsection{Experiments Setup}
We conduct a comprehensive evaluation of our proposed method across three demanding tasks of 3D reconstruction, video depth estimation, and camera pose estimation. Initially, we leverage the longest contiguous sequences from established public datasets, benchmarking \mymethod{} against leading long-term streaming baselines, namely CUT3R~\cite{2025cut3r} and TTT3R~\cite{2025ttt3r}. 
Our method is a training-free optimization designed to overcome the memory bottlenecks inherent in long-sequence reconstruction. We implement and evaluate this approach on the StreamVGGT~\cite{2025streamVGGT}, concentrating our analysis on long-sequence scenarios where the benefits of our memory-efficient design are most pronounced. On shorter sequences, where the baseline's GPU memory is not exceeded, performance differences are negligible (see Sec.~\ref{sec:Discussion}). Subsequently, we introduce and evaluate on our novel large-scale \textbf{Long3D} benchmark to probe stability across extensively prolonged inputs. All experiments were executed on a single NVIDIA A100 GPU.
 
\subsection{3D Reconstruction}
\noindent \textbf{Evaluation on 7-Scenes and NRGBD Datasets.} 
Following the previous work~\cite{2025cut3r}, we evaluate scene-level 3d reconstruction on 7-Scenes~\cite{20137scenes} and NRGBD~\cite{2022NeuralRGBD} datasets. 
But unlike the evaluation of extremely sparse-view reconstruction protocol before, for the long-term streaming, we sampled images with stride = 2 in each sequence and use the first 300 to 500 images as input like TTT3R~\cite{2025ttt3r}. As shown in Tab.~\ref{tab:mv_recon_long}, the offline method VGGT~\cite{2025vggt} and the online method StreamVGGT~\cite{2025streamVGGT} fail on long sequences input as the memory constraints.
As for the other runnable online method, while TTT3R maintains robust performance on the 7-Scenes dataset, its reconstruction capability on NRGBD degrades significantly as the number of input frames increases.
Despite affording greater robustness on varied datasets, Point3R's explicit pointer mechanism~\cite{2025point3r} suffers from perpetually increasing memory usage, rendering it incompatible with long-sequence reconstruction (evidenced in \cite{2025ttt3r}).
Our method \mymethod{} exhibits minimal temporal error accumulation as the sequence length increases, allowing it to consistently maintain the state-of-art reconstruction accuracy.
Concurrently, its strong performance across varied datasets highlights its high robustness.
\begin{figure*}[t]
    \centering
    \includegraphics[width=\textwidth]{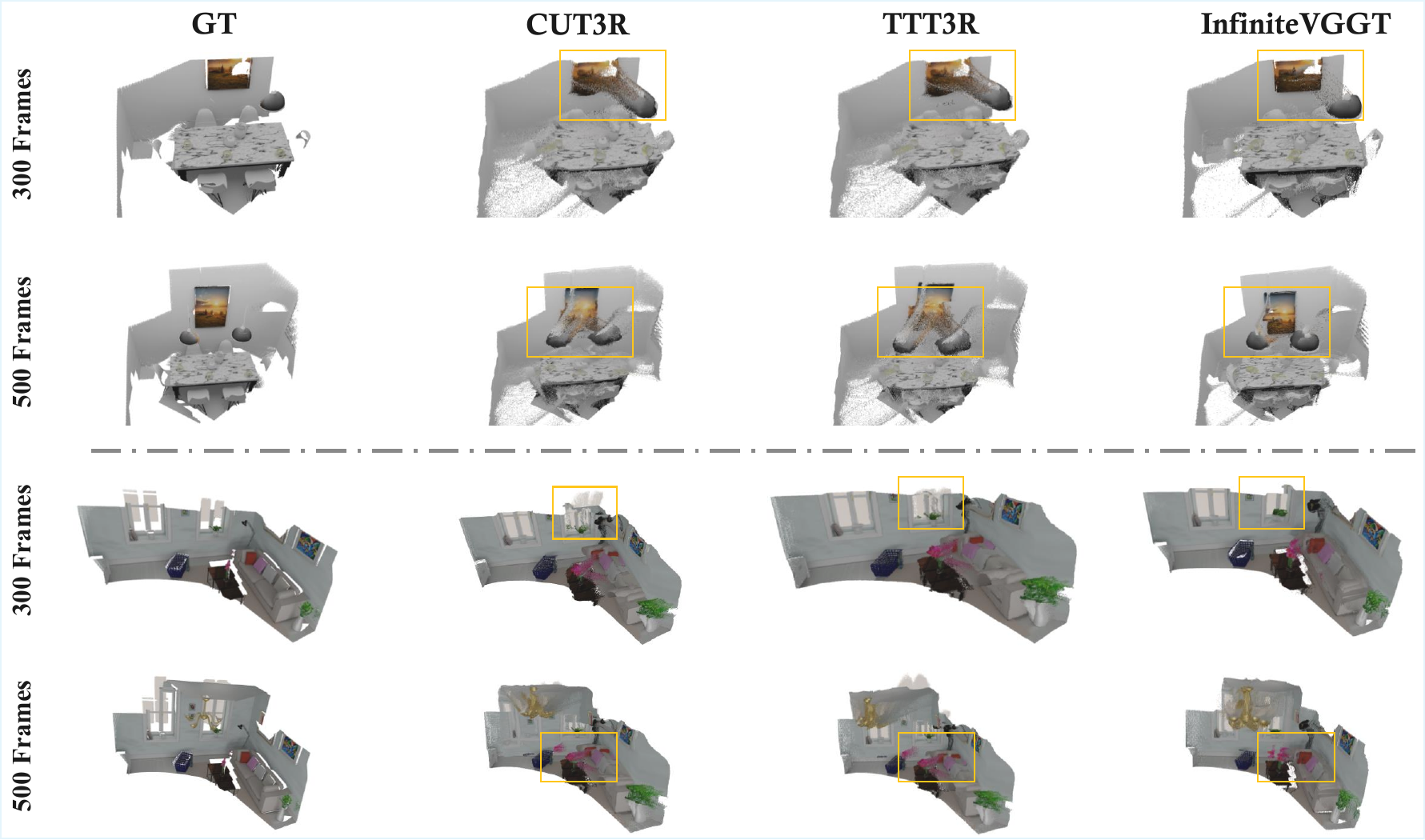}
    
    \caption{\textbf{Qualitative Results of 3D Reconstruction.}}
    \label{fig:qualitative}
    \vspace{-1.5em}
\end{figure*}
\begin{table}[t]
\centering
\caption{
    \textbf{3D Reconstruction Results on Long3D.}
}
\vspace{-1em}
\resizebox{\columnwidth}{!}{
\begin{tabular}{lccccccccc}
    \toprule[0.17em]
    {\multirow{2}{*}{\textbf{Method}}} &
    {\multirow{2}{*}{\textbf{Scene/Input}}} &
    \multicolumn{2}{c}{Acc. $\downarrow$} &
    \multicolumn{2}{c}{Comp. $\downarrow$} &
    \multicolumn{2}{c}{NC $\uparrow$} &
    {\multirow{2}{*}{CD $\downarrow$}} &\\
    \cmidrule(r){3-4} \cmidrule(r){5-6} \cmidrule(r){7-8}
    & &
    Mean & Med. &
    Mean & Med. &
    Mean & Med. & \\
    \midrule[0.08em]
    CUT3R~\cite{2025cut3r} & \multirow{3}{*}{\makecell{\textit{Classroom} \\ 2128}} & 0.496 & 0.374 & 0.085 & 0.036 & 0.520 & 0.525 & 0.291\\
    TTT3R~\cite{2025ttt3r} && 0.396 & 0.319 & 0.081 & 0.035 &0.530 & 0.540 & 0.239 \\
    \cellcolor{lightblue}\textbf{\NickName \mymethod{}} && \cellcolor{lightblue}\textbf{0.357} & \cellcolor{lightblue}\textbf{0.298} & \cellcolor{lightblue}\textbf{0.057} & \cellcolor{lightblue}\textbf{0.033} & \cellcolor{lightblue}\textbf{0.576} & \cellcolor{lightblue}\textbf{0.612} & \cellcolor{lightblue}\textbf{0.207} \\
    \midrule[0.08em]
    CUT3R~\cite{2025cut3r} &\multirow{3}{*}{\makecell{\textit{Dormitory} \\ 4208}}& 1.800 & 1.372  & 0.404 & 0.090  & 0.501  &0.495 & 1.102\\
    TTT3R~\cite{2025ttt3r} && 1.965 & 1.749 & \textbf{0.329} & 0.100 & 0.515 & 0.509 & 1.147\\
    \cellcolor{lightblue}\textbf{\NickName \mymethod{}} && \cellcolor{lightblue}\textbf{1.438} & 
    \cellcolor{lightblue}\textbf{1.159} & 
    \cellcolor{lightblue}0.575 & 
    \cellcolor{lightblue}\textbf{0.089} & 
    \cellcolor{lightblue}\textbf{0.526} & 
    \cellcolor{lightblue}\textbf{0.538} & 
    \cellcolor{lightblue}\textbf{1.007}\\
    \midrule[0.08em]
    CUT3R~\cite{2025cut3r} &\multirow{3}{*}{\makecell{\textit{Library} \\ 4726}}& 1.907 & 1.437 & \textbf{0.193} & 0.079 & 0.504 & 0.507& 1.050\\
    TTT3R~\cite{2025ttt3r} && 2.175 & 1.484 & 0.430 & 0.095 & 0.494 &0.481 & 1.303 \\
    \cellcolor{lightblue}\textbf{\NickName \mymethod{}} && \cellcolor{lightblue}\textbf{1.121} & 
    \cellcolor{lightblue}\textbf{0.821} & 
    \cellcolor{lightblue}0.571 & 
    \cellcolor{lightblue}\textbf{0.077} & 
    \cellcolor{lightblue}\textbf{0.508} & 
    \cellcolor{lightblue}\textbf{0.514} & 
    \cellcolor{lightblue}\textbf{0.846} \\
    \midrule[0.08em]
    CUT3R~\cite{2025cut3r} &\multirow{3}{*}{\makecell{\textit{Badminton Court} \\ 6067}}& 2.489 & 2.432 & 5.802&5.071 & 0.495 &0.483 &4.146 \\
    TTT3R~\cite{2025ttt3r} &&2.791  &2.392  &3.160  &2.673  & 0.509 & 0.502 & 2.975\\
     \cellcolor{lightblue}\textbf{\NickName \mymethod{}} &&  \cellcolor{lightblue}\textbf{1.843} &  \cellcolor{lightblue}\textbf{1.555} & \cellcolor{lightblue}\textbf{1.854} & \cellcolor{lightblue}\textbf{0.816} & \cellcolor{lightblue}\textbf{0.510}  &  \cellcolor{lightblue}\textbf{0.509} &  \cellcolor{lightblue}\textbf{1.848}\\
    \midrule[0.08em]
    CUT3R~\cite{2025cut3r} & \multirow{3}{*}{\makecell{\textit{Academic Building} \\ 9545}} & 8.062 & 5.650 & \textbf{0.673} & 0.251 & 0.496 & 0.491 & 4.638\\
    TTT3R~\cite{2025ttt3r} &&7.710 & 5.793 & 6.192 & 5.159 & \textbf{0.513} & \textbf{0.519} & 6.951\\
    \cellcolor{lightblue}\textbf{\NickName \mymethod{}} && \cellcolor{lightblue}\textbf{5.733} & \cellcolor{lightblue}\textbf{4.603} & \cellcolor{lightblue}1.206 & \cellcolor{lightblue}\textbf{0.251} & 
    \cellcolor{lightblue}0.495 & 
    \cellcolor{lightblue}0.490 & 
    \cellcolor{lightblue}\textbf{3.470}\\
    \bottomrule[0.17em]

\end{tabular}
}
\vspace{-1em}
\label{tab:mv_recon_Long3D}
\end{table}

\noindent \textbf{Evaluation on Long3D Benchmark.} 
As Sec.~\ref{sec:Benchmark} states, we evaluated our method alongside other models capable of processing extended-length inputs on sequences of approximately 2,000, 4,500, 6000 and nearly 10,000 frames on Long3D dataset.
The results demonstrate that our approach achieves robust performance across diverse scenes and varying sequence lengths, outperforming existing models like CUT3R~\cite{2025cut3r} and TTT3R~\cite{2025ttt3r} on most metrics.
More importantly, although temporal drift inevitably accumulates with increasing input frames, our method effectively limits this error propagation compared to baselines.
However, we observed that our method underperforms on the mean of Comp. metric compared to these baselines. 
We identify this as a key area for optimization in our future work.

\subsection{Video Depth Estimation}
\noindent \textbf{Evaluation on Bonn Datasets.}
Video depth estimation evaluates per-frame depth quality and inter-frame depth consistency.
Since most existing datasets only contain a limited number of continuous frames, to show the long-term performance, we evaluate \mymethod{} on the longest available continuous sequences from Bonn~\cite{2019bonn}.
Specifically, we select continuous sequences ranging from 200 to 500 frames, beginning after the initial 30 frames like TTT3R.
As shown in Tab.~\ref{tab:videodepth}, the performance of \mymethod{} is benchmarked against CUT3R~\cite{2025cut3r} and TTT3R~\cite{2025ttt3r}, showing the effectiveness of our method.

\begin{table}[t]
    \centering
    \caption{
        \textbf{Video Depth Estimation on Bonn~\cite{2019bonn}.} 
    }
    \vspace{-1em}
    \resizebox{1.0\columnwidth}!{
    \begin{tabular}{lccccccc}
        \toprule[0.17em]
        {\multirow{2}{*}{\textbf{Method}}} &
        {\multirow{2}{*}{\textbf{Input}}} &
        \multicolumn{2}{c}{\textbf{Bonn}} &\\
        \cmidrule(r){3-4}
        & &
        Abs Rel $\downarrow$ & $\delta<1.25$ $\uparrow$&\\
        \midrule
        VGGT \textit{(Offline)}~\cite{2025vggt} & \multirow{5}{*}{\makecell{\textit{200}}}  & \textcolor{lightgray}{\textit{OOM}} & \textcolor{lightgray}{\textit{OOM}}\\ 
        StreamVGGT~\cite{2025streamVGGT} & & \textcolor{lightgray}{\textit{OOM}} & \textcolor{lightgray}{\textit{OOM}}\\
        CUT3R~\cite{2025cut3r} & & 0.072 & 0.947\\
        Point3R~\cite{2025point3r} & & 0.069 & 0.954\\
        TTT3R~\cite{2025ttt3r} & & 0.068 & 0.953\\
        \rowcolor{lightblue} \textbf{\NickName \mymethod{}} & & \textbf{0.063} & \textbf{0.964}\\
        \midrule
        VGGT \textit{(Offline)}~\cite{2025vggt} & \multirow{5}{*}{\makecell{\textit{300}}}  & \textcolor{lightgray}{\textit{OOM}} & \textcolor{lightgray}{\textit{OOM}}\\
        StreamVGGT~\cite{2025streamVGGT} & & \textcolor{lightgray}{\textit{OOM}} & \textcolor{lightgray}{\textit{OOM}}\\
        CUT3R~\cite{2025cut3r} & & 0.089 & 0.938\\
        Point3R~\cite{2025point3r} & & 0.081 & 0.946\\
        TTT3R~\cite{2025ttt3r} & & 0.079 & 0.949\\
        \rowcolor{lightblue} \textbf{\NickName \mymethod{}} & & \textbf{0.072} & \textbf{0.958}\\
        \midrule
        VGGT \textit{(Offline)}~\cite{2025vggt} & \multirow{5}{*}{\makecell{\textit{400}}}  & \textcolor{lightgray}{\textit{OOM}} & \textcolor{lightgray}{\textit{OOM}}\\
        StreamVGGT~\cite{2025streamVGGT} & & \textcolor{lightgray}{\textit{OOM}} & \textcolor{lightgray}{\textit{OOM}}\\
        CUT3R~\cite{2025cut3r} & & 0.090 & 0.934\\
        Point3R~\cite{2025point3r} & & 0.081 & 0.945\\
        TTT3R~\cite{2025ttt3r} & & 0.078 & 0.951\\
        \rowcolor{lightblue} \textbf{\NickName \mymethod{}} & & \textbf{0.070} & \textbf{0.958}\\
        \midrule
        VGGT \textit{(Offline)}~\cite{2025vggt} & \multirow{5}{*}{\makecell{\textit{500}}}  & \textcolor{lightgray}{\textit{OOM}} & \textcolor{lightgray}{\textit{OOM}}\\
        StreamVGGT~\cite{2025streamVGGT} & & \textcolor{lightgray}{\textit{OOM}} & \textcolor{lightgray}{\textit{OOM}}\\
        CUT3R~\cite{2025cut3r} & & 0.084 & 0.939\\
        Point3R~\cite{2025point3r} & & 0.081 & 0.946\\
        TTT3R~\cite{2025ttt3r} & & 0.076 & 0.953\\
        \rowcolor{lightblue} \textbf{\NickName \mymethod{}} & & \textbf{0.069} & \textbf{0.960}\\
        \bottomrule[0.17em]
    \end{tabular}
    }
    \vspace{-1em}
    \label{tab:videodepth}
\end{table}

\subsection{Ablation Study}
\noindent \textbf{Crucial Token Selection Policy.}
We conduct a comparative analysis of attention weight-based and cosine similarity-based token selection policies on the 7-Scenes dataset~\cite{20137scenes}.
In addition to evaluating reconstruction quality via chamfer distance (CD), and normal consistency (NC) metrics, chamfer distance is computed as the average of accuracy and completeness.
We also profile the per-frame inference time and peak GPU memory consumption. 
As summarized in Tab.~\ref{tab:attn and cossim}, the cosine similarity-based approach yields more accurate point cloud reconstruction. 
Moreover,  standard attention weight-based methods can introduce an additional 120ms of inference latency per frame, while our model's compatibility with FlashAttention~\cite{2022FlashAttention} mitigates this bottleneck, enabling significantly faster inference speeds and a reduced peak GPU memory footprint.

\noindent \textbf{Initial Budget Per-head.}
We further ablate the budget $B^{(l,h)}$ using the 7-Scenes dataset, comparing the results for 300 and 500 input with a stride of 2, where $B^{(l,h)}$ denotes the initial maximum storage capacity for tokens per head in each layer.
As shown in Tab.~\ref{tab:initial budget}, a smaller token storage budget $B^{(l,h)}$ significantly degrades the reconstruction quality, while this impact diminishes as the budget increases, eventually becoming negligible.

\noindent \textbf{Layer-wise Allocation Mechanism.}
To demonstrate the effectiveness of our layer-wise allocation mechanism for token selection, we conducted an ablation study on the 7-Scenes. The input frames are 500.
Given an initial budget $B_{10000}^{(l,h)}$, we compare maintaining a uniform, fixed storage limit across all layers against our dynamic layer-wise allocation scheme.
As shown in Tab.~\ref{tab:layer-wise allocate}, dynamically allocating the budget across layers further improves the resulting point cloud accuracy and normal consistency.

\noindent \textbf{Anchor Frame.}
The VGGT~\cite{2025vggt} architecture fundamentally depends on the first frame as a global reference and establishes the canonical coordinate system for the entire sequence.
Given this pivotal role, we posit that applying token pruning to the initial state could lead to irreversible information loss. 
Therefore, we design a strategy where the tokens of the first frame are fully retained as an anchor frame, effectively bypassing the diversity-based selection mechanism applied to subsequent frames.
To validate the necessity of this design, we conduct an ablation study on the 7-Scenes~\cite{20137scenes} dataset, using 300-frame inputs sampled with a stride of 2, to assess the impact of this anchor frame strategy on 3D reconstruction accuracy. 
As evidenced by the results in Tab.~\ref{tab:anchor frame}, preserving the complete reference information of the first frame prevents error accumulation and leads to a significant improvement in reconstruction quality.
\begin{table}[t]
    \centering
    \caption{
        \textbf{Ablation on Attention and Cosine Similarity Method.} 
    }
    \vspace{-1em}
    \resizebox{1.0\columnwidth}!{
    \begin{tabular}{lcccc}
        \toprule[0.17em]
        Method &
        CD $\downarrow$ &
        NC $\uparrow$ &
        Time (s) $\downarrow$ &
        Peak Memory (GB) $\downarrow$\\
        \midrule
        Attention weight & 0.036 & 0.567 & 0.288 & 17.30\\
        Cosine similarity & \textbf{0.032} & \textbf{0.570} & \textbf{0.168} & \textbf{14.49}\\
        \bottomrule[0.17em]
    \end{tabular}
    }
    \vspace{-1em}
    \label{tab:attn and cossim}
\end{table}
\begin{table}[t]
    \centering
    \caption{
        \textbf{Ablation Study on Initial Budget Per-Head.} 
    }
    \vspace{-1em}
    \resizebox{1.0\columnwidth}!{
    \begin{tabular}{lccccc} 
        \toprule[0.17em]
        \multirow{2}{*}{\diagbox{Initial Budget}{Input}}
        & \multicolumn{2}{c}{300} & \multicolumn{2}{c}{500} \\
        \cmidrule(r){2-3} \cmidrule(r){4-5} 
        & CD $\downarrow$ & NC $\uparrow$ & CD $\downarrow$ & NC $\uparrow$ \\
        \midrule
        $B^{l,h}_{10000}$ & 0.062 & 0.565 & 0.075 & 0.555 \\ 
        $B^{l,h}_{25000}$ & 0.032 & 0.570 & 0.033 & 0.560 \\
        $B^{l,h}_{50000}$ & 0.032 & 0.570 & 0.031 & 0.562 \\
        \bottomrule[0.17em]
    \end{tabular}
    }
    \vspace{-1em}
    \label{tab:initial budget}
\end{table}
\begin{table}[t]
\centering
\caption{
    \textbf{Ablation Study on Layer-wise Allocation Mechanism.}
}
\vspace{-1em}
\resizebox{\columnwidth}{!}{
\begin{tabular}{lcccccccccccccc}
    \toprule[0.17em]
    {\multirow{2}{*}{Method}} &
    {\multirow{2}{*}{Input}} &
    \multicolumn{2}{c}{Acc. $\downarrow$} &
    \multicolumn{2}{c}{Comp. $\downarrow$} &
    \multicolumn{2}{c}{NC $\uparrow$} & \\
    \cmidrule(r){3-4} \cmidrule(r){5-6} \cmidrule(r){7-8}
    & &
    Mean & Med. &
    Mean & Med. &
    Mean & Med. & \\
    \midrule[0.08em]
    w/o layer-wise allocation & {\multirow{2}{*}{500}} & 0.098 & 0.058 & 0.057 & 0.008 & 0.554 & 0.582 \\
    w/ layer-wise allocation && \textbf{0.093} & \textbf{0.053} & \textbf{0.056} & \textbf{0.008} & \textbf{0.555} & \textbf{0.583}\\
    \bottomrule[0.17em]
\end{tabular}
}
\vspace{-1em}
\label{tab:layer-wise allocate}
\end{table}
\begin{table}[t]
\centering
\caption{
    \textbf{Ablation Study on Anchor Frame Mechanism.}
}
\vspace{-1em}
\resizebox{\columnwidth}{!}{
\begin{tabular}{lccccccccccccc}
    \toprule[0.17em]
    {\multirow{2}{*}{Method}} &
    \multicolumn{2}{c}{Acc. $\downarrow$} &
    \multicolumn{2}{c}{Comp. $\downarrow$} &
    \multicolumn{2}{c}{NC $\uparrow$} & \\
    \cmidrule(r){2-3} \cmidrule(r){4-5} \cmidrule(r){6-7}
    &
    Mean & Med. &
    Mean & Med. &
    Mean & Med. & \\
    \midrule[0.08em]
    w/o anchor frame &  0.047 & 0.020 & 0.027 & 0.006 & 0.570 & 0.606 \\
    w/  anchor frame & 0.040 & 0.015 & 0.025 & 0.006 & 0.570 & 0.607\\
    \bottomrule[0.17em]
\end{tabular}
}
\vspace{-1em}
\label{tab:anchor frame}
\end{table}
\section{Discussion}
\label{sec:Discussion}
\begin{figure}[tb!]
    \centering
    \includegraphics[width=0.99\linewidth]{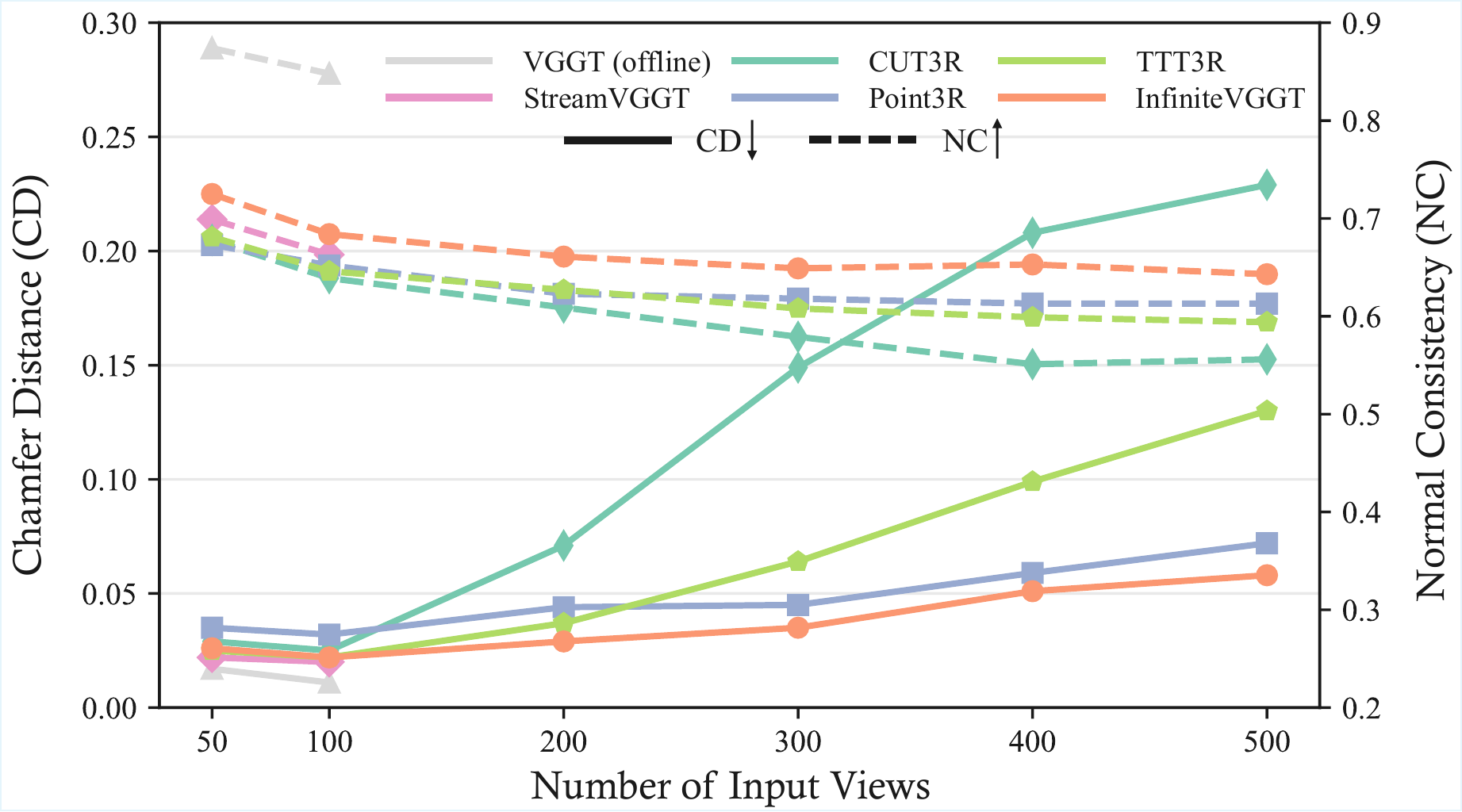}
    \caption{\textbf{Comparison of 3D Reconstruction.} CD and NC metrics on NRGBD~\cite{2022NeuralRGBD} dataset.}
    \label{fig:discussion}
    \vspace{-15pt}
\end{figure}

The primary objective of this work is to enable online, infinite-horizon 3D geometry estimation for streaming scenes through a novel rolling memory mechanism. Given that our method, \mymethod{}, is a training-free modification of StreamVGGT~\cite{2025streamVGGT}, its performance on shorter sequences relative to the baseline warrants clarification. We therefore begin by confirming that our approach achieves comparable performance in these less demanding scenarios. On input sequences from 50 to 100 frames, which is a range where the baseline operates without memory constraints, our comparison of CD and NC metrics reveals negligible performance differences. As shown in Fig.~\ref{fig:discussion}, \mymethod{} even achieves a slight precision advantage in the NC metric. This advancements arise from our diversity-aware rolling memory mechanism, which refines the model's historical context. By preserving a more diverse set of information from early stages, the mechanism enhances robustness against noisy data encountered as the sequence grows. It also prevents the memory from being saturated by redundant subsequent inputs. For long sequences, these benefits become critical. \mymethod{} not only resolves the out-of-memory (OOM) errors that plague the baseline but also curtails the accumulation of temporal error. In line with our primary goal of addressing the challenges of long-sequence reconstruction, our evaluation is therefore concentrated on these demanding scenarios.

\section{Conclusion}
\label{sec:Conclusion}

We present \mymethod{}, a novel rolling memory paradigm for streaming 3D geometry understanding that mitigates the trade-off between unbounded memory growth and long-term drift. Our training-free strategy achieves this by identifying memory redundancy via key cosine similarity and applying an adaptive, layer-wise budget allocation. This mechanism, fully compatible with FlashAttention, ensures bounded memory and computational efficiency for online streaming over infinite-horizon sequences. As a result, \mymethod{} surpasses existing explicit- and implicit-state methods in reconstruction accuracy and robustness. We also introduce the Long3D benchmark to support rigorous evaluation of extended-sequence performance.

{
    \small
    \bibliographystyle{ieeenat_fullname}
    \bibliography{main}
}

\end{document}